\documentclass[10pt,twocolumn,letterpaper]{article}

\usepackage{cvpr}              

\usepackage[dvipsnames]{xcolor}

\newlength\paramargin
\newlength\figmargin

\newlength\secmargin
\newlength\figcapmargin
\newlength\tabcapmargin

\newcommand{\mpage}[2]
{
\begin{minipage}{#1\linewidth}\centering
#2
\end{minipage}
}

\newcommand{\figcaption}[2]
{
\caption{
\textbf{#1.}  
#2            
}
}

\newcommand{\figref}[1]{Figure~\ref{fig:#1}} 
\newcommand{\tabref}[1]{Table~\ref{tab:#1}}

\long\def\ignorethis#1{}

\newbox\jsavebox%

\definecolor{cvprblue}{rgb}{0.21,0.49,0.74}
\usepackage[pagebackref,breaklinks,colorlinks,citecolor=cvprblue]{hyperref}

\usepackage{multicol}
\usepackage{multirow}
\usepackage[accsupp]{axessibility} 
\usepackage{amssymb}
\usepackage{pifont}
\usepackage{appendix}
\newcommand{\cmark}{\ding{51}}%
\newcommand{\xmark}{\ding{55}}%
\usepackage{graphicx}
\def\paperID{8636} 
\def\confName{CVPR}
\def\confYear{2024}

\title{Do You Remember? Dense Video Captioning with Cross-Modal \\Memory Retrieval}

\author{Minkuk Kim$^{1}$, Hyeon Bae Kim$^{1}$, Jinyoung Moon$^{2}$, Jinwoo Choi$^{1,}$\textsuperscript{*}, Seong Tae Kim$^{1,}$\thanks{Dr. J. Choi and Dr. S. T. Kim are corresponding authors.}\\
$^1$Kyung Hee University, Republic of Korea \\ 
$^2$Electronics and Telecommunications Research Institute (ETRI), Republic of Korea \\
}

\begin{document}
\maketitle
\begin{abstract}
There has been significant attention to the research on dense video captioning, which aims to automatically localize and caption all events within untrimmed video. Several studies introduce methods by designing dense video captioning as a multitasking problem of event localization and event captioning to consider inter-task relations. However, addressing both tasks using only visual input is challenging due to the lack of semantic content. In this study, we address this by proposing a novel framework inspired by the cognitive information processing of humans. Our model utilizes external memory to incorporate prior knowledge. The memory retrieval method is proposed with cross-modal video-to-text matching. To effectively incorporate retrieved text features, the versatile encoder and the decoder with visual and textual cross-attention modules are designed. Comparative experiments have been conducted to show the effectiveness of the proposed method on ActivityNet Captions and YouCook2 datasets. Experimental results show promising performance of our model without extensive pretraining from a large video dataset. Our code is available at \url{https://github.com/ailab-kyunghee/CM2_DVC}.
\end{abstract}    
\section{Introduction}
\label{sec:intro}

With the increasing demand for video understanding and multimodal analysis, the field of video captioning is growing rapidly.
The task of conventional video captioning involves generating precise descriptions for trimmed video segments and several studies show successful results~\cite{lin2022swinbert,luo2020univl,seo2022end,7984828,pei2019memory,qi2019sports,wang2018reconstruction,chen2017video,pan2016jointly,rohrbach2013translating,venugopalan2014translating,venugopalan2015sequence}. However, it faces considerable challenges when applied to dense video captioning. Dense video captioning aims to localize important event segments (i.e., to find
event boundaries) from untrimmed videos and describe the event segment (i.e., what happens in the event) with natural language. 
For achieving high-performance dense video captioning, it is important to properly model inter-task interactions between event localization and caption generation. 

\begin{figure}[t]
    \centering
    \includegraphics[width=0.99\linewidth]{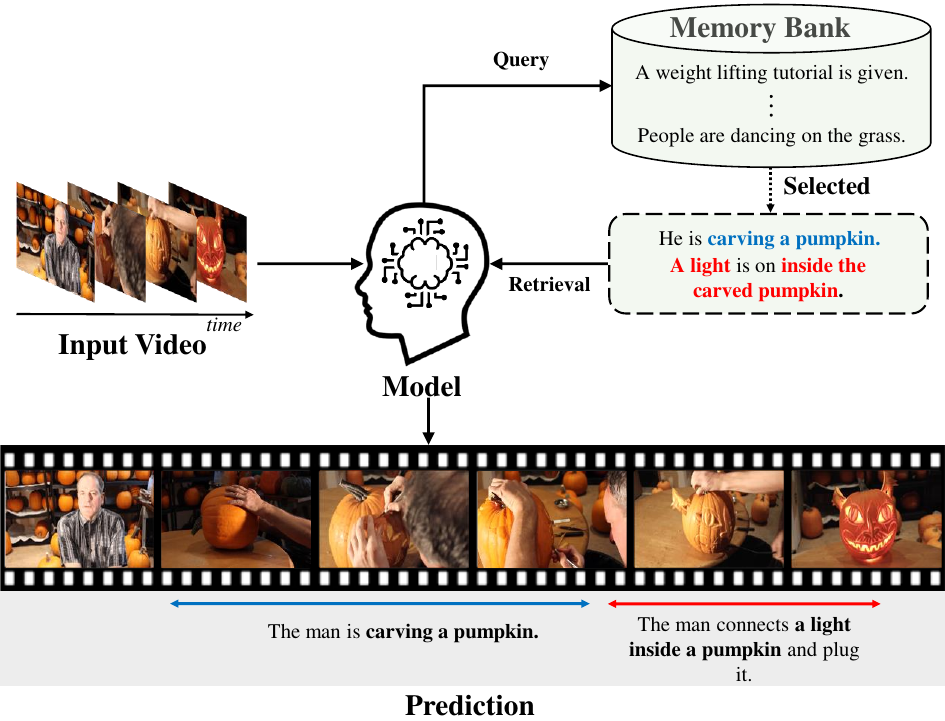}
    \caption{Conceptual figure of the proposed cross-modal memory-based dense video captioning (CM$^2$). Our method can search for relevant clues from an external memory bank to provide precise descriptions and localization for untrimmed video.}
    \label{fig:example}
\end{figure}

Recent studies in vision and language learning have shown impressive results in cross-modal correlation tasks \cite{radford2021learning,saharia2022photorealistic,li2022blip}. However, connecting natural language and video is still challenging due to the difficulties in modeling spatiotemporal information \cite{hayes2022mugen}. Video-and-language learning requires complex model architectures, specialized training protocols, and large computational costs~\cite{cheng2023vindlu}.
Even dense video captioning requires connecting untrimmed videos and natural language to localize events and describe them~\cite{krishna2017dense,wang2021end,yang2023vid2seq}. 

This study is motivated by the observation of how humans recognize and describe scenes. Humans are capable of identifying important events and describing them by recalling relevant memories based on cues they have observed. In cognitive information processing, this processing is called cued recall~\cite{allan1997event,rugg1998neural}.
By recalling relevant memories, humans can describe the scenes with human-understandable natural language.

To verify the feasibility of our idea, we have conducted a preliminary experiment. To measure the usefulness of text clues from external memory, we search the relevant information by using the ground truth caption of the query video, which is the ideal case where we can achieve in the external memory. In the real-world condition, we could not use text query and video features will be used as a query. As shown in Table~\ref{tab:retrieval_w/o}, the performance of the dense video captioning could be significantly improved (CIDEr of 183.95 is achievable with Oracle retrieval on the ground truth event segment in YouCook2 dataset \cite{zhou2018towards}).

Following this insight, we devise a new dense video captioning framework, named Cross-Modal Memory-based dense video captioning (CM$^2$). Our model can recall relevant events from external memory to improve the generation quality of captions in dense video captioning as shown in Figure \ref{fig:example}. To mimic the human's process, an external memory is designed based on prior knowledge which is extracted from training data. Then, the proposed model extracts potential event candidates from given untrimmed videos and retrieves relevant information from the external memory to provide the model with diverse and semantic information. By incorporating the retrieved memory into visual features, our method further introduces a versatile encoder and decoder structure. The encoded features are aggregated by using visual cross-attention and textual cross-attention in a versatile transformer decoder, which helps the model learn inter-task interactions from visual and text clues. Our main contributions can be summarized as:
\begin{itemize}
    \item Inspired by the human cognitive process, we introduce a new dense video captioning method with cross-modal retrieval from external memory. To the best of our knowledge, this is the first study that uses cross-modal retrieval from external memory for dense video captioning. By retrieving relevant text clues from the memory, the proposed model could elaborately localize and describe important events in a more fluent and natural way.

    \item To effectively leverage multi-modal features, we propose a versatile encoder-decoder structure with a visual cross-attention and a textual cross-attention. Our model could effectively learn cross-modal correlation and model inter-task interactions for improving dense video captioning.

    \item 
    Comprehensive experiments have been conducted on ActivityNet Captions \cite{krishna2017dense} and YouCook2 \cite{zhou2018towards} datasets to verify the effectiveness of memory retrieval in dense video captioning. Our model also achieves comparable performance without pretraining on large video datasets. 
\end{itemize}

\section{Related Work}
\label{sec:RelatedWork}

\subsection{Desne Video Captioning} Dense video captioning is a multi-task problem that combines two sub-tasks: Event localization and event captioning. Krishna \textit{et al}. \cite{krishna2017dense} introduced a dense video captioning model by first generating proposals and then using an attention-based LSTM to generate captions, following the "localize-then-describe" strategy. Subsequent studies \cite{iashin2020multi,iashin2020better,yang2018hierarchical,wang2018bidirectional,wang2020event} aimed to produce more precise and informative captions within this strategy. However, two-stage approaches have major limitations, as they do not jointly train event localization and event captioning, resulting in less attention to inter-task interactions.

To address the aforementioned limitations, recent studies propose joint training of two sub-tasks \cite{Deng_2021_CVPR,wang2021end,yang2023vid2seq,chadha2020iperceive,chen2021towards,li2018jointly,mun2019streamlined,rahman2019watch,shen2017weakly,shi2019dense,wang2018bidirectional,zhou2018end}. Deng \textit{et al}. \cite{Deng_2021_CVPR} initially generate a paragraph for a given video and then utilize it for grounding.
Wang \textit{et al}. \cite{wang2021end} define dense video captioning as a parallel set prediction task and propose an end-to-end method for event localization and event captioning, using only visual input to solve the two sub-tasks. 
Yang \textit{et al}. \cite{yang2023vid2seq} make use of transcribed speech for multi-modal inputs, predicting both time tokens and caption tokens as a single sequence. For the pretraining of the model, an additional YT-Temporal-1B dataset which contains 18 million narrated videos collected from
YouTube is used. 

However, training high-quality dense video captioning models without pretraining from a large number of videos still remains very challenging. 
Our study presents a novel approach to exploit prior knowledge to enhance the quality of dense video captioning.

\subsection{Retrieval-Augmented Generation} The retrieval-augmented approach is often used in language generation tasks. Lewis \textit{et al}.\cite{lewis2020retrieval} propose retrieval-augmented generation, which combines pre-trained parametric and external non-parametric memory to effectively leverage pre-trained model knowledge. Some works \cite{ramos2021retrieval,ramos2023retrieval,sarto2022retrieval,xu2019unified,zhao2020image} in image captioning also employ this external datastore approach. Similar to ours, Sarto \textit{et al}.\cite{sarto2022retrieval} and Ramos \textit{et al}.\cite{ramos2023retrieval} propose an approach to train a retrieval-augmented image captioning model by processing encoded retrieved captions through cross-attention.
Recent studies also show retrieval augmented generation in the context of video captioning \cite{zhang2021open,chen2023retrieval,10183355}. They propose to improve video captioning by incorporating external knowledge, such as video-related training corpus \cite{10183355} and memory-augmented encoder-decoder structure \cite{zhang2021open}. They reference the retrieved text obtained from memory in the word prediction distribution of the captioning decoder. 

In this study, our model references retrieved information throughout all layers of the decoder with cross-attention. While they only concentrate on enhancing word prediction for video captioning, our method adopts a structure that utilizes the retrieved text as semantic information, benefiting both event localization and event captioning.
Note that, retrieval-augmented generation has been largely unexplored in dense video captioning. Previous studies that use the retrieval-augmented generation approach in the image and short video captioning only utilize retrieved textual information for improving caption quality.
In this study, we present a new structure to exploit the retrieval of text clues for generating dense captions and localizing events from untrimmed videos.

\section{Method}
Our goal is to improve event-level localization and event captioning from untrimmed video by exploiting prior knowledge. For this, we introduce a new framework (CM$^2$) which is designed with cross-modal memory retrieval. CM$^2$ could search relevant information by segment-level video features and retrieve text features from external memory in a video-to-text cross-modal manner (Section~\ref{sec:MemRetrieval}). 

\begin{figure*}[t]
\centering
\mpage{1.0}{
    \includegraphics[width=0.8\textwidth]{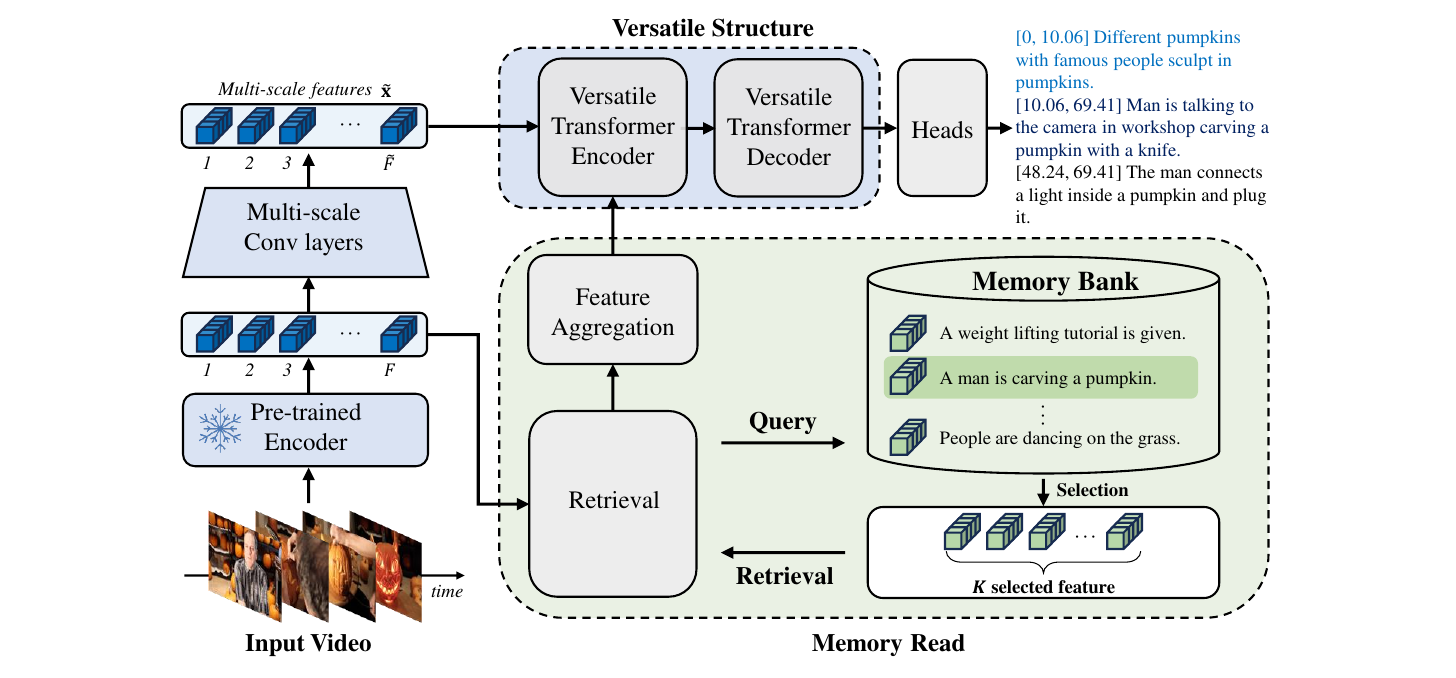}
}
\mpage{1.0}{
    (a) Overall Architecture.
}
\vfill
\mpage{0.49}{
    \includegraphics[width=0.92\textwidth]{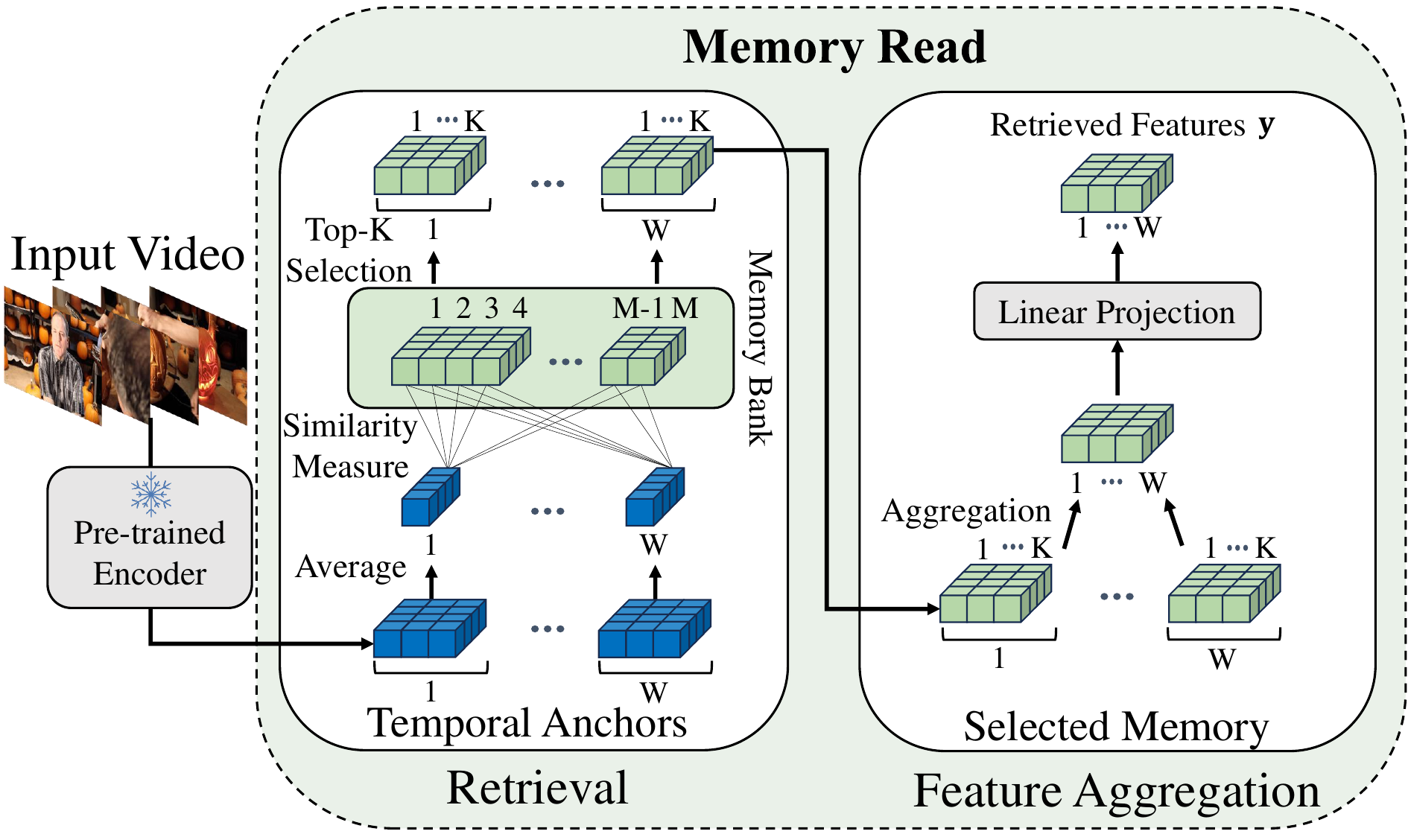}
}
\hfill
\mpage{0.49}{
    \includegraphics[width=0.92\textwidth]{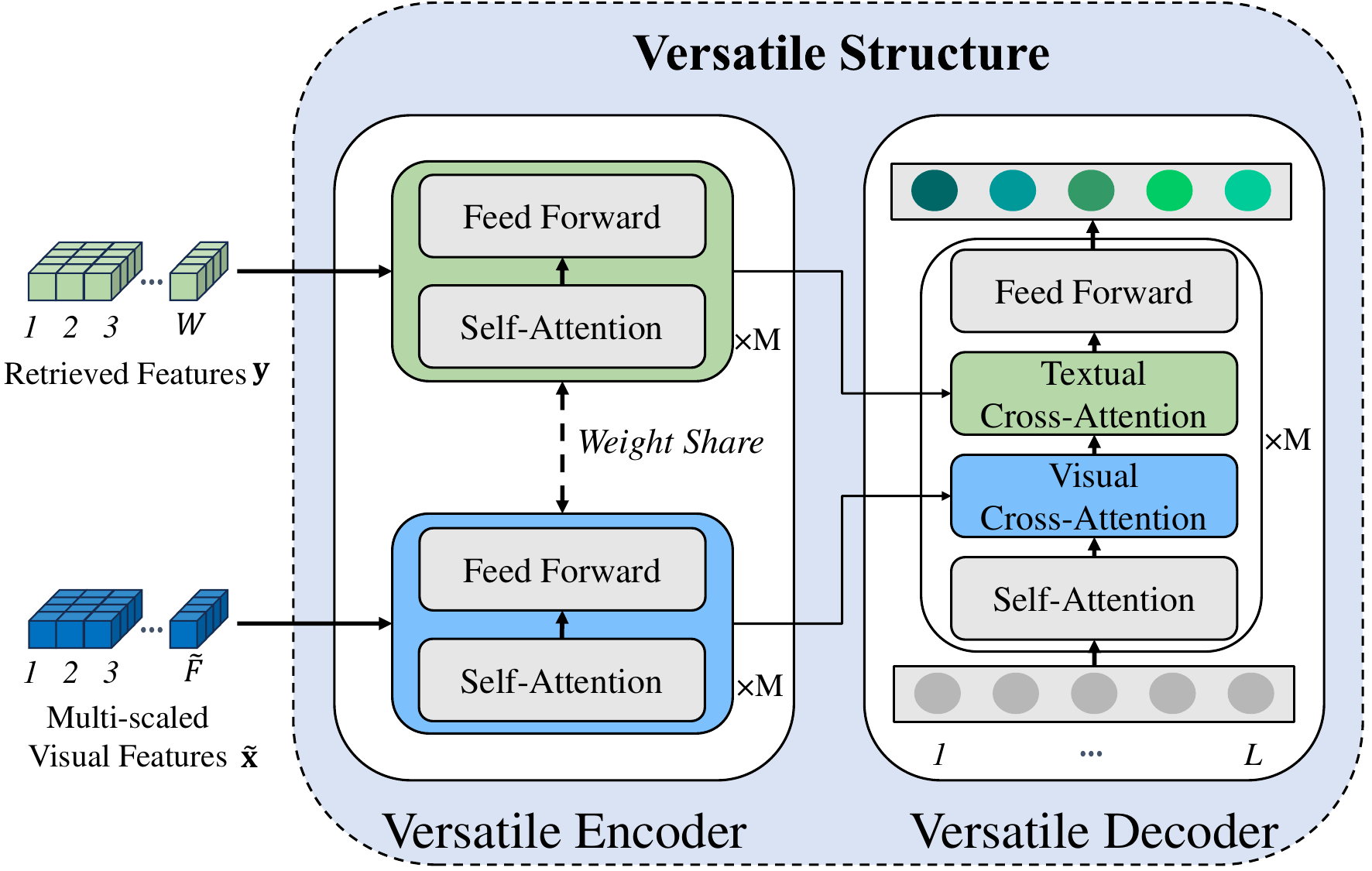}
}

\mpage{0.49}{
    (b) Memory Read Module.
}
\hfill
\mpage{0.49}{
    (c) Versatile Encoder-Decoder Module.
}
\vspace{-0.2cm}
\figcaption{Overview of CM$^2$}{We approach the dense video captioning task in a memory-retrieval-augmented caption generation manner. We show the overall architecture in (a). We conduct video-to-text cross-modal retrieval using input video features obtained through a pre-trained encoder. As illustrated in (b), we generate segment-level $W$ temporal anchors from the input video features. Then we measure similarities between the anchors and the text features stored in a memory to obtain $W$ retrieved features through aggregation. As illustrated in (c), we encode the multi-scale video features $\tilde{\textbf{x}}$ and retrieved features $\textbf{}$ using a versatile transformer encoder. Each encoded feature vector undergoes the corresponding cross-attention layers to obtain refined event queries. Finally, we obtain the set of start time, end time, and caption by passing the event queries through a head.
}
\label{fig:overview}
\end{figure*}

Furthermore, to ensure that the model efficiently leverages the retrieved semantic information for both localization and captioning tasks, we design a versatile encoder-decoder architecture and a modal-level cross-attention method (Section~\ref{sec:versatilED}). As illustrated in Figure \ref{fig:overview}, our model takes input video frames and extracts video frame features \(\textbf{x}=\{x_i\}^F_{i=1}\) and retrieved text features \(\textbf{y}=\{y_j\}^W_{j=1}\) where $F$ and $W$ denote the number of frames in the given video and the number of retrieved text features, respectively. For the given input video, the model generates segment and caption pairs \(\{(t^s_n,t^e_n,S_n)\}^{N}_{n=1}\) where $N$ denotes the number of events detected by our method and $t^s_n$ and $t^e_n$ denote the start and the end timestamp of $n$-th event. $S_n$ denotes the generated captions for $n$-th event segment. Details of dense event prediction will be introduced in Section~\ref{sec:desnseprediction}.  

\subsection{Memory Retrieval}
\label{sec:MemRetrieval}
\subsubsection{Memory Construction} 
To store high-quality semantic information as prior knowledge in the memory, we first construct an explicit external memory bank by encoding sentence-level features. The sentences are collected from the training data of the in-domain target dataset ~\cite{zhou2018towards,krishna2017dense} by taking into account the semantic distributions appropriate to the query videos. For example, the captions in AcitivityNet Caption training set are used for constructing external memory in experiments
on AcitivityNet Captions in this study. For segment-level video-to-text retrieval, we define a memory unit in a sentence level that corresponds to the event clip instead of whole paragraphs from an untrimmed video in the dataset. 
 
For segment-level video-to-text embedding, we adopt pre-trained CLIP Vit-L/14 \cite{dosovitskiy2020image,radford2021learning} which shows promising alignment ability by mapping image and text to the shared feature space. 
For storing semantic information of captions at a sentence level, we tokenize the captions of the event segment using the CLIP tokenizer, ensuring padding to match the maximum token number of ground truth captions. Subsequently, all tokenized caption sentences are encoded by a CLIP text encoder, and the resulting sentence-level embeddings are stored in the external memory bank.

\subsubsection{Segment-level Retrieval}
Untrimmed videos could consist of multiple events, each containing distinct semantic information. As both sub-tasks of dense video captioning operate at the event level, it is crucial to design an appropriate retrieval method that considers segment-level semantic information. We propose a novel cross-modal memory-based dense video captioning (CM$^2$), designed to take into account the semantic information of the segment that can potentially include events. By utilizing image-to-text retrieval strategies with CLIP~\cite{dosovitskiy2020image,radford2021learning} and temporal anchors, our method ensures the incorporation of semantic details from dense events. The proposed approach involves two key steps: segment-level retrieval and feature aggregation as shown in Figure \ref{fig:overview} (b).

In segment-level retrieval, to acquire semantic information related to events within the input video, we divide the input video into $W$ temporal anchors. For frame-level visual feature extraction, we adopt CLIP ViT-L/14. To obtain the representative information contained in each anchor, we compress the temporal dimension at each anchor through averaging, yielding segment-level visual features. Then, for each anchor, the segment-level visual feature is used as a query for retrieving relevant information from the external memory. For finding relevant information, the similarity between the segment-level visual feature and CLIP text features in the memory is calculated (In this study, cosine similarity between two feature vectors is used as a similarity metric). 
Based on the similarity scores, $K$ sentence features are retrieved for each anchor, which results in a selected memory feature set for $j$-th anchor as $\textbf{m}^j=\{m^{j}_1, ..., m^{j}_K\}$.

Next, we perform feature aggregation to summarize useful information from $K$ retrieved sentence features $\textbf{m}^j$ associated with each anchor in the selected memory. The average pooling over the $K$ sequences is conducted in each anchor. Finally, we obtain the retrieved text features $\{y_j\}^W_{j=1}$.

\subsection{Versatile Encoder-Decoder}
\label{sec:versatilED}
In this section, we describe how we build a structure to incorporate visual features and retrieved text features for event localization and event captioning. We generate event query features with well-incorporated temporal information using an encoder-decoder structure based on the deformable transformer~\cite{zhu2020deformable}, as in~\cite{wang2021end}. However, our approach differs from~\cite{wang2021end} as our model incorporates not only visual features but also retrieved text features for making positive effects in both captioning and localization. To achieve this, we propose a versatile encoder-decoder structure that effectively uses retrieved text features and visual features.

\noindent \textbf{Feature Encoding.} First, we sample the frame-level features extracted by the pre-trained CLIP ViT-L/14 with 1 FPS to a fixed frame number as  \(\textbf{x}=\{x_i\}^F_{i=1}\) for batch processing. Then, we added $L$ temporal convolutional layers for the multi-scaling processing of video frame features. The multi-scale convolutional layers output multi-scale visual features as \(\tilde{\textbf{x}}=\{\tilde{x}_i\}^{\tilde{F}}_{i=1}\).

\noindent \textbf{Versatile encoder.} CM$^2$ enhances the interplay between visual and text modalities while preserving their original information, achieved through the use of versatile weight-shared encoders. These weight-shared encoders, illustrated in \figref{overview} (c), are employed to process each modality feature. The versatile encoder is designed with $M$ blocks where each block consists of feedforward and self-attention layers. By employing weight-shared encoders, the visual and text modality features undergo training in a shared embedding space, fostering potential cross-modality connections. Furthermore, since each modality process is processed separately by the weight-shared encoder, it could effectively retain distinctive modality-specific information. The visual encoder takes a sequence of multi-scale frame features \(\tilde{\textbf{x}}=\{\tilde{x}_i\}^{\tilde{F}}_{i=1}\) as input and generates encoded visual features as output. Simultaneously, the same versatile encoder processes a set of retrieved text features \(\textbf{y}=\{y_j\}^W_{j=1}\), producing $W$ encoded text features.

\begin{table*}[t]
\caption{\textbf{Effect of memory retrieval in ActivityNet Captions and Youcook2.} No retrieval refers to a case where the model is forwarded without any retrieval. Oracle methods are implemented to measure the upper bound which could be achieved by the retrieval with an ideal query to the memory bank. The captions retrieved from the memory by ground truth captions of query video are directly used as the output of the model.
}
\centering
\scalebox{0.89}{
\begin{tabular}{@{}l|cccc|cccc@{}}
\toprule
\multirow{2}{*}{Retrieval Type} &
\multicolumn{4}{|c}{ActivityNet} & \multicolumn{4}{|c}{YouCook2} \\ 
& CIDEr &METEOR & BLEU4&SODA$\_c$ & CIDEr  & METEOR & BLEU4 & SODA$\_c$ \\
\midrule
No Retrieval & 31.24 & 8.03 &  2.15 &6.01 & 23.67 & 5.30 & 1.17&4.77  \\
Proposed Retrieval (Ours)& 33.01 & 8.55&2.38 &6.18  & 31.66 & 6.08 & 1.63&5.34 \\
Oracle w/o GT proposal & 40.24 & 9.43 & 2.88 &6.96  & 53.55 & 9.18& 3.49 & 6.81  \\
Oracle w/ GT proposal & 84.47 & 15.69 &5.86& 12.41  &183.95 & 23.53 & 13.05&25.51  \\
\bottomrule
\end{tabular} 
}
\label{tab:retrieval_w/o}
\end{table*}

\noindent \textbf{Versatile decoder.} Through the versatile decoder, we design learnable embeddings, event queries \(\textbf{q}=\{q_l\}^L_{l=1}\), to include temporally and semantically rich information. When video and text modalities are given, a single cross-attention is insufficient to generate the necessary representations for the two sub-tasks. Therefore, CM$^2$ separates the visual cross-attention layer from the textual cross-attention layer, as described in \figref{overview} (c). We aim for each modality to handle tasks related to temporal and semantic information processing separately. In visual cross-attention, considering the cross-attention between encoded visual features and event queries enhances the temporal information of event queries. In textual cross-attention, considering the cross-attention between encoded text features and event queries enriches the semantic information of event queries. The output of the versatile decoder produces event queries \(\tilde{\textbf{q}}=\{q_l\}^L_{l=1}\) with both temporal and semantic information.

\subsection{Dense Event Prediction}
\label{sec:desnseprediction}
\textbf{Parallel Heads.}
CM$^2$ employs a parallel decoding structure with sub-task heads for a given event query $\tilde{q}_l$. Our approach includes three parallel heads: a localization head, a captioning head, and an event counter. 
\newline \textbf{Localization Head.} The localization head is implemented by a multi-layer perceptron to predict the box prediction, including the center and length of the ground-truth segment, for a given event query. Additionally, it conducts binary classification to predict the foreground confidence of each event query. Finally, the localization head outputs a set of tuples ${(t^s_l, t^e_l, c_l)}^{L}_{l=1}$, where each tuple represents the start time $t^e_l$, end time $t^e_l$, and localization confidence $c_l$ of $l$-th event segment, respectively.
\newline \textbf{Captioning Head.} For the captioning head, we employ the deformable soft attention LSTM which uses the soft attention around the reference points, enhancing word generation performance. ~\cite{wang2021end}. For the input of the captioning head, we utilize attention feature $a_{l,s}$, event query $\tilde{q}_l$, and the previous word $w_{l,s-1}$ to predict the next word. As the sentence progresses, the captioning head generates the entire sentence $\textbf{S}_l={w_{l,1},...,w_{l,S}}$, where $S$ represents the length of the sentence.
\newline \textbf{Event Counter.} The event counter predicts the appropriate number of events in the video. To achieve this, it compresses essential information from the event query $\tilde{q}_l$ through a max-pooling layer and a fully-connected layer. It predicts a vector $r_{len}$ representing a specific number of events. During inference, the predicted event count is selected by $N=argmax(r_{len})$. Finally, the N predicted sets $\{(t^s_n,t^e_n,S_n)\}^{N}_{n=1}$ are determined by the Hungarian algorithm \cite{carion2020end}, using a matching cost $C=L_{cls}+\alpha L_{loc}$ with generalized IOU loss and focal loss. The focal loss $L_{cls}$ is computed between the predicted classification score and the ground-truth label. The generalized IOU loss $L_{loc}$ measures the predicted segment against the ground-truth segment.
\newline \textbf{Training and Inference.} During training, we train CM$^2$ using four losses: $L_{loc}$, $L_{cls}$, $L_{count}$, and $L_{cap}$. $L_{count}$ represents the cross-entropy between the predicted count number distribution and the ground truth. $L_{cap}$ is the cross-entropy between the predicted word probability and the ground truth. The total loss is defined as follows:
\begin{equation}
L_T = L_{cls}+\lambda_{loc}L_{loc}+\lambda_{count}L_{count}+\lambda_{cap}L_{cap}
\end{equation}
During inference, given visual input $x$ and retrieved text input $y$, our model predicts $N$ sets of predictions $\{(t^s_n,t^e_n,S_n)\}^{N}_{n=1}$.
For both training and inference, we conducted retrieval using the same external memory bank.
\section{Experiments}
To verify the effectiveness of our method, comparative experiments have been conducted. First, Section~\ref{subsection:1} introduces the experimental setting used in this study. Section~\ref{subsection:EMR} shows the effectiveness of memory retrieval in dense video captioning. Section~\ref{subsection:SOTA} shows the comparison with state-of-the-art methods. Section~\ref{subsection:abl} shows ablation studies for our model to validate the effectiveness of each component. Qualitative results of our method and discussion are followed.

\subsection{Experimental Settings}
\label{subsection:1}
\textbf{Dataset.} We employed two dense video captioning benchmark datasets, namely ActivityNet Captions~\cite{krishna2017dense} and YouCook2 \cite{zhou2018towards}, for training and evaluation. ActivityNet Captions consists of 20k untrimmed videos of diverse human activities. On average, each video spans 120s and is annotated with 3.7 temporally localized sentences. For training, validation, and testing, we follow the standard split of videos. YouCook2 consists of 2k untrimmed cooking procedure videos, with an average duration of 320s per video and 7.7 temporally localized sentences per annotation.  We followed the standard split for training, validation, and testing videos. Notably, we use approximately 7\% fewer videos than the original count, as we use those accessible on YouTube.

\noindent \textbf{Evaluation Metrics.} We evaluated our method for two sub-tasks in dense video captioning. By using ActivityNet Challenge official evaluation tool~\cite{wang2020dense}, we evaluated generated captions using the metrics CIDEr~\cite{vedantam2015cider}, BLEU4~\cite{papineni2002bleu}, and METEOR~\cite{banerjee2005meteor}, which calculate matched pairs between generated captions and ground truth across IOU thresholds of {0.3, 0.5, 0.7, 0.9}. Additionally, for measuring storytelling ability, we employed SODA\_c~\cite{fujita2020soda}. For event localization, we measured average precision, average recall, and F1 score, which represents the harmonic mean of precision and recall. These scores are averaged over IOU thresholds of {0.3, 0.5, 0.7, 0.9}.

\noindent \textbf{Implementation Details.} For both datasets, we extract video frames at a rate of 1 frame per second and then subsample or pad the sequence of frames to achieve a total of $F$ frames, where we set $F = 100$ in ActivityNet Captions and $F = 200$ in YouCook2. We employ a two-layer deformable transformer with multiscale deformable attention spanning four levels. The number of event queries is set to 10 for ActivityNet Captions and 100 for YouCook2, respectively. 
In this study, the balancing hyperparameters of $\alpha$, $\lambda_{loc}$, $\lambda_{count}$, and $\lambda_{cap}$ are set to 2, 2, 1, and 1, respectively. 
The number of anchors is empirically set to 10 for ActivityNet Captions and 50 for YouCook2. In retrieval, We set the anchor number of 50, with k set to 80 for each anchor. Therefore, we utilize 4000 retrieved text features. During training, the ground truth of the corresponding input video was excluded from the memory bank.

\begin{table}[t!]
    \caption{\textbf{Performance of Event Captioning in ActivityNet Captions.}
        Bold means the highest score. Underline means 2nd score. \# PT denotes the number of videos used for pretraining. {$^{\dagger}$} denotes results reproduced from official implementation in our environment. 
    }
    \centering
    \resizebox{1.0\columnwidth}{!}{
        \begin{tabular}{@{}l|l|c|cccc@{}}
        \toprule
        Method & Backbone & \# PT & CIDEr & METEOR & BLEU4 & SODA$\_c$ \\
        \midrule
        Vid2Seq~\cite{yang2023vid2seq} & CLIP & 15M  & \underline{30.10} & \underline{8.50} & - & 5.80   \\
        \midrule
        MT~\cite{zhou2018end} & TSN & - & 6.10 & 3.20 & 0.30 & - \\
        ECHR~\cite{wang2020event} & C3D & - & 14.70 & 7.20 & 1.82 & 3.20     \\
        UEDVC~\cite{zhang2022unifying} & C3D & - & - & - & - & 5.5 \\
        PDVC{$^{\dagger}$}~\cite{wang2021end} & CLIP & - & 29.97 & 8.06 &\underline{2.21} & \underline{5.92} \\
        \textbf{Ours} & CLIP & - & \textbf{33.01} & \textbf{8.55} & \textbf{2.38} & \textbf{6.18}  \\
        \bottomrule
        \end{tabular}
    }
\label{tab:anet-dvc}
\end{table}
\subsection{Effect of Memory Retrieval}
\label{subsection:EMR}

To assess the effectiveness of memory retrieval, comparative experiments have been conducted. Four different memory retrieval approaches are implemented as shown in \tabref{retrieval_w/o}. No retrieval refers to a method where the model is forwarded without any retrieval. 
Oracle retrieval is implemented to measure the upper bound of memory retrieval. The captions are retrieved from the external memory based on the similarity with ground truth captions of query video. The retrieved captions are directly used as an output of the model without model forwarding. For Oracle without GT proposal, matched the retrieved text to the event segments predicted by our model. For Oracle with GT proposal match retrieved text to ground truth event segments for measuring the performance.

As shown in the table, the proposed retrieval method achieves higher scores compared with the model without retrieval. This is mainly due to the reason that retrieved text features could provide semantically useful features to the model, which helps the model exploit visual and text relations. 

When we evaluate the performance with Oracle, even when we used the same event segments as our model, by using the retrieved text without the model forward, we observed a large enhancement in captioning performance. Moreover, when we match retrieved text to ground truth event segments, the performance is significantly improved. 

These results show huge potential for retrieval-based dense video captioning. In this study, we use clip visual features from Vit-L/14 and average the features from each anchor are aggregated by averaging them. In other words, video-to-text matching is implemented by projecting video features to image-to-text feature space. Some important video features might be lost during this process. In the future, according to the advances in video modeling and video-to-text matching, our method which uses retrieval from external memory for dense video captioning could be further improved.

\subsection{Comparison with State-of-the-art-Methods}
\label{subsection:SOTA}

\begin{table}[t!]
    \caption{\textbf{Performance of Event Captioning in YouCook2.}
        Bold means the highest score. Underline means 2nd score. \# PT denotes the number of videos used for pretraining. {$^{\dagger}$} denotes results reproduced from official implementation in our environment.
    }
    \centering
    \resizebox{1.0\columnwidth}{!}{        
        \begin{tabular}{@{}l|l|c|cccc@{}}
        \toprule
        Method & Backbone & \# PT & CIDEr & METEOR & BLEU4 & SODA$\_c$\\
        \midrule
        Vid2Seq~\cite{yang2023vid2seq} & CLIP & 1M  &  \textbf{47.10} & \textbf{9.30} & - & \textbf{7.90} \\
        \midrule
        MT~\cite{zhou2018end} & TSN & -  & 9.30 & 5.00 & 1.15 & - \\
        ECHR~\cite{wang2020event} & C3D & -  & - & 3.82 & - & -    \\
        E2ESG~\cite{zhu2022end} & C3D & - &25.00 & 3.50 & - & - \\
        PDVC{$^{\dagger}$}~\cite{wang2021end} & CLIP & - & 29.69 & 5.56 & 1.40 & 4.92  \\
        \textbf{Ours} & CLIP & - & \underline{31.66} & \underline{6.08} & \textbf{1.63} & \underline{5.34} \\
        \bottomrule
        \end{tabular}
    }
\label{tab:yc2-dvc}
\end{table}

In \tabref{anet-dvc} and \tabref{yc2-dvc}, we compare our method with state-of-the-art dense video captioning approaches~\cite{zhou2018end, wang2020event, wang2021end, zhang2022unifying, zhu2022end, yang2023vid2seq} on both YouCook2 and ActivityNet Captions datasets. 
As shown in \tabref{anet-dvc}, our method achieves the best scores over four metrics of CIDEr, METEOR, BLEU4, and SODA$\_c$. Even the method could achieve higher scores compared with \cite{yang2023vid2seq} which leverages an additional 15 million videos for pretraining.
In YouCook2 dataset, Vid2seq~\cite{yang2023vid2seq} which uses extra 1 million videos for pretraining achieves the best performance. Our method achieves comparable performance on YouCook2 without using extra videos. By using prior knowledge from external memory, our method could improve the quality of caption generation. 

\begin{table}[t!]
\caption{\textbf{Performance of Event Localization in ActivityNet Captions and YouCook2.} Bold means the highest score. Underline means 2nd score. PT denotes pretraining from the additional video datasets. {$^{\dagger}$} denotes results reproduced from official implementation in our environment. All methods used the CLIP as the backbone. }
\centering
\resizebox{1.0\columnwidth}{!}{
    \begin{tabular}{@{}l|c|ccc|ccc@{}}
    \toprule
    \multirow{2}{*}{Method}&\multirow{2}{*}{PT} & \multicolumn{3}{|c}{ActivityNet Captions} & \multicolumn{3}{|c}{YouCook2} \\
    &&F1& Recall & Precision & F1&Recall & Precision \\
    \midrule
    Vid2Seq~\cite{yang2023vid2seq}&\cmark  & 53.29 & 52.70 & 53.90 & \underline{27.84}  & \textbf{27.90} & 27.80  \\
    \midrule
    PDVC{$^{\dagger}$}~\cite{wang2021end} &\xmark& \underline{54.78} & \underline{53.27} & \underline{56.38}  & 26.81 & 22.89 & \underline{32.37}  \\
    \textbf{Ours}&\xmark & \textbf{55.21} & \textbf{53.71} & \textbf{56.81} & \textbf{28.43}& \underline{24.76} & \textbf{33.38} \\
    \bottomrule
    \end{tabular}
}
\label{tab:loc}
\end{table}

We also compare the localization ability of our method. \tabref{loc} shows the comparison of our model with other models that use CLIP features as a visual feature in YouCook2 and ActivityNet Captions datasets. As shown in the table, our method achieves the best scores in ActivityNet Captions in both precision and recall. Also, in YouCook2 dataset, our method achieves the best precision and second recall scores, which results in the best F1 score. 
Our memory retrieval approach not only improves caption generation but also helps the model to localize event boundaries by providing semantic cues that can be exploited during inter-task interactions.

\begin{figure*}[t]
    \centering
    \includegraphics[width=0.99\linewidth]{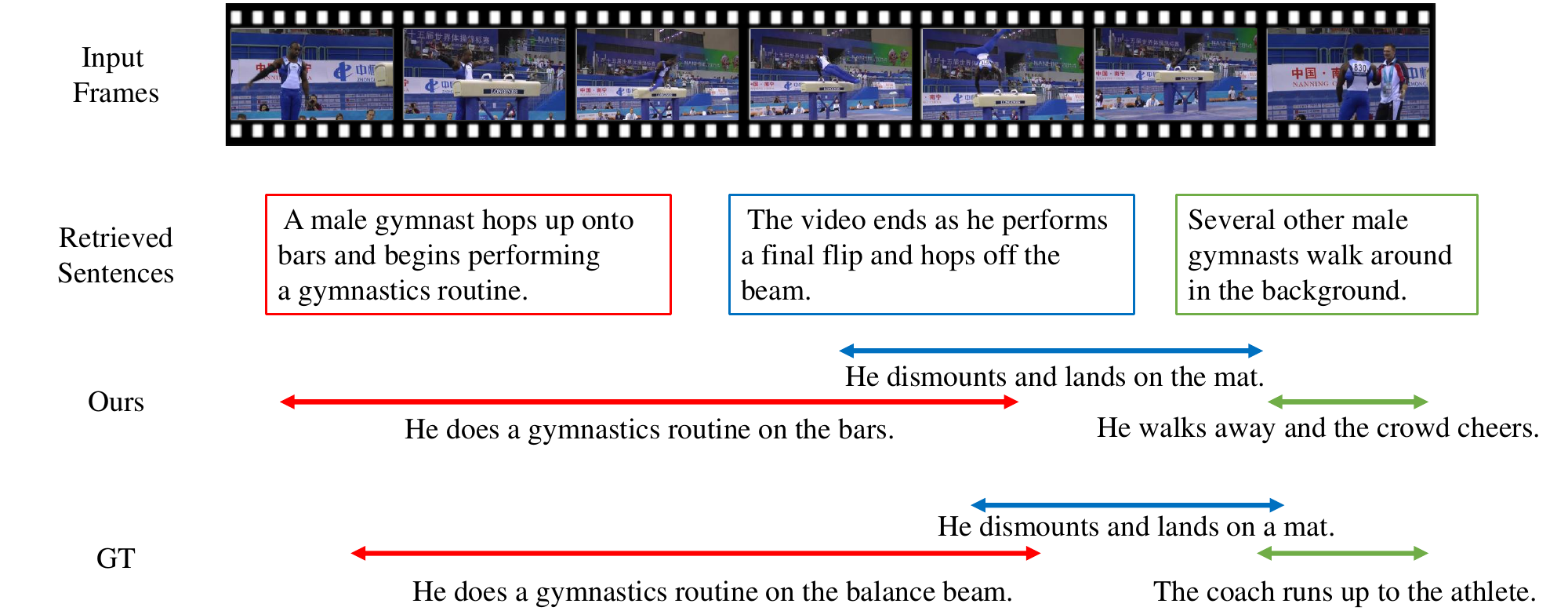}
    \caption{\textbf{Example of dense video captioning predictions with ours on ActivityNet Captions Validation set.} We show a comparison with the ground truth. Retrieved sentences are example results from retrieval that have the highest semantic similarity to the corresponding segments of input frames. Each retrieved sentence is utilized in our model's predictions for the segments with the corresponding color.}
    \label{fig:qualitative}
\end{figure*}

\subsection{Ablation Studies}
\label{subsection:abl}

\begin{table}[t!]
\caption{\textbf{Ablation study to verify the effect of structure component for incorporating retrieved features.} WS denotes a weight sharing for the versatile encoder. SE denotes the case where we encode textual and visual features, separately. Without SE is implemented by concatenating textual and visual features and passing through a single encoder. TCA denotes the use of textual cross-attention where the model uses additional cross-attention for encoded text features. Without TCA is implemented by concatenating visual and text features and passing through a single cross-attention. The performance is measured in YouCook2.}
\centering
\resizebox{1\columnwidth}{!}{
    \begin{tabular}{@{}ccc|ccccc@{}}
    \toprule
    WS & SE & TCA & CIDEr & METEOR &BLEU4&  SODA$\_c$ &F1 \\
    \midrule
    \xmark & \xmark & \xmark& 29.49 & 5.65 & 1.34 &\underline{5.26} 
     &\underline{27.66} \\
    \xmark & \xmark& \cmark & 28.40 & 5.56 & 1.37& 4.74 & 25.14 \\
    \xmark&\cmark & \xmark & 30.86  & 5.61 & 1.40 & 5.14 & 27.06 \\
    \xmark &\cmark & \cmark & \underline{30.94}  &  \underline{5.71} & \textbf{1.65} & 5.07 & 27.11 \\
    \cmark &\cmark & \cmark & \textbf{31.66} & \textbf{6.08}&\underline{1.63}  & \textbf{5.34} & \textbf{28.43} \\
    \bottomrule
    \end{tabular}
}
\vspace{-0.2cm}
\label{tab:ret_structure}
\end{table}

Our method aims to leverage the retrieved segment-level text features as semantic information for improving dense video captioning. In this section, we present ablation studies for the component that is designed to incorporate retrieved features from the memory. We design a versatile encoder structure where the encoder processes retrieved text features and visual features. 
In other words, one encoder is shared between two modalities, and the model is trained to process both modalities. \tabref{ret_structure} shows the ablation study results. 
It is observed that the use of a weight-shared versatile encoder structure could improve the model performance. The cases where cross-modal information (i.e., visual and textual features) is processed by the separate encoder (with SE) achieve higher performance compared with the model without the separate encoder in which the two features are concatenated before entering the transformer encoder. Also, weight sharing for the encoder is better than having two separate encoders for each modality. These results indicate that it is important to encode visual and textual features separately by preserving own information. However, the model could learn the interconnection between textual and visual features by training the encoder in a versatile manner.

Furthermore, we also compare the presence of textual cross-attention. We compare cases where our model is designed with separate textual and visual cross-attention with the cases where the model is implemented by using combined cross-attention where the textual and visual features are concatenated before being put into the decoder. As shown in \tabref{ret_structure}, the performance is increased with separate textual cross-attention. This is mainly due to the reason that the decoder could incorporate visual and textual features by specialized cross-attention explicitly. 

\subsection{Discussion}
\subsubsection{Qualitative Examples}
Figure~\ref{fig:qualitative} shows predicted examples of our approach. It can be observed that memory retrieval effectively references meaningful and helpful sentences from memory, obtained through segment-level video-text retrieval for the given video. As a result, our method generates relatively accurate event boundaries and captions.  The semantic information obtained from memory through retrieval assists in semantic predictions during caption generation. More examples are provided in Supplementary Material.

\subsubsection{Effect of Anchor Number for Retrieval}
\begin{table}[t!]
\caption{\textbf{Effect of anchor number for retrieval in YouCook2.} \# Anchor denotes the number of anchors. The performance is measured by changing the number of anchors.
}
\centering
\resizebox{0.92\columnwidth}{!}{
    \begin{tabular}{@{}c|ccccc@{}}
    \toprule
    \multirow{1}{*}{\# Anchor} & CIDEr & METEOR & BLEU4 &SODA$\_c$& F1 \\
    \midrule
    1 & 27.97 & 5.54 & 1.39 &5.14 & \underline{28.10} \\
    10 & 31.36 & 5.75 & \underline{1.63} &5.17 & 27.33  \\
    30 & 28.41 & \underline{6.02} &1.43& 5.08 & 26.87 \\
    50 & \underline{31.66} & \textbf{6.08} & \underline{1.63} &\textbf{5.34} & \textbf{28.43} \\
    70 & \textbf{32.73} & 5.83 & \textbf{1.66} &\underline{5.28} & 27.55 \\
    90 & 29.88 & 5.57 & 1.43 &\textbf{5.34} & 27.63 \\
    \bottomrule
    \end{tabular}
}
\vspace{-0.2cm}
\label{tab:anchor_num}
\end{table}

We explore the effect of the number of temporal anchors generated during memory retrieval. The number of temporal anchors is related to the basic unit for giving a query to the memory bank and it also attributes to the number of retrieved features. \tabref{anchor_num} shows the performance by changing the anchor number in YouCook2 dataset. When the anchor number is set to 1, the untrimmed video information is averaged to a single visual feature for querying to the memory bank. This approach could not exploit fine-grained details for retrieving the semantic text cues. As we increase the number of anchors, the fine-grained details can be captured for querying to the memory bank, which improves the performance of dense video captioning model. However, when an excessive number of features are retrieved, noisy features contribute to the degradation of performance.
It is observed that the anchor number of 50 consistently yields outstanding performance in both event localization and caption generation in YouCook2 dataset. 

\begin{table}[t!]
\caption{\textbf{Effect of the number of selected features in YouCook2.} \#SF denotes the number of retrieved features from the memory bank. The performance is measured by changing the number of retrieved text features per anchor. 
}
\centering
\scalebox{0.85}{
\begin{tabular}{@{}c|ccccc@{}}
\toprule
\multirow{1}{*}{\#SF} &  CIDEr & METEOR & BLEU4 & SODA$\_c$& F1 \\
\midrule
1 &19.76 & 4.36 & 0.65& 4.79 &26.79\\
20 & 30.22 & 5.64 & 1.62 & 5.20  &27.33  \\
40 & 31.25& 5.73 & \textbf{1.79}&5.29  &28.10 \\
60 & 31.24 & 5.77 & \underline{1.63}& 5.26 & \textbf{28.58} \\
80 & \underline{31.66} & \textbf{6.08}&\underline{1.63} & \textbf{5.34} & \underline{28.43} \\
100 & \textbf{32.07} & \underline{5.81} & 1.58&\underline{5.32} & 27.86 \\
\bottomrule
\end{tabular}
}
\vspace{-0.2cm}
\label{tab:topk}
\end{table}

\subsubsection{Effect of Number of Retrieved Features for Each Anchor}

We also investigate the effect of the number of retrieved features per anchor on the performance. 
When we set the number of retrieved features per temporal anchor to 1, it means we only consider the text from memory that has the highest similarity to the visual feature of the temporal anchor. \tabref{topk} shows the results according to the number of retrieved features per anchor. 
As we increase the number of retrieved features per anchor, the memory read could provide stable and robust semantic information to the model. When the number of retrieved features per anchor is set to 80, our method consistently achieves good performance in both sub-tasks in YouCook2 dataset. 
However, with a too large number, the noisy features could be retrieved because we retrieved the features with the similarity in descending order, which could degrade the performance.

\section{Conclusion}
In this study, we introduced a novel approach to dense video captioning inspired by the human cognitive process of scene understanding. Leveraging cross-modal retrieval from external memory, CM$^2$ demonstrated a significant improvement in both event localization and caption generation. Through comprehensive experiments on ActivityNet Captions and YouCook2 datasets, we validated the effectiveness of our memory retrieval approach. Notably, CM$^2$ achieved competitive results without the need for pre-training on a large number of video data, highlighting its efficiency. We believe that our work opens avenues for future study in dense video captioning and encourages the exploration of memory-augmented models for improving video understanding and captioning.
\section*{Acknowledgements}
This work was supported in part by the Institute of Information and Communications Technology Planning and Evaluation (IITP) Grant funded by the Korea Government (MSIT) under Grant 2020-0-00004 (Development of Provisional Intelligence Based on Long-term Visual Memory Network), Grant 2022-0-00078, Grant IITP-2024-RS-2023-00258649, Grant 2021-0-02068, Grant RS-2022-00155911, by the National Research Foundation of Korea (NRF) Grant funded by the Korea Government (MSIT) under Grant 2021R1G1A1094990, and by Center for Applied Research in Artificial Intelligence (CARAI) grant funded by DAPA and ADD (UD230017TD).
{
    \small
    \bibliographystyle{ieeenat_fullname}
    \bibliography{main}

\begin{thebibliography}{56}
\providecommand{\natexlab}[1]{#1}
\providecommand{\url}[1]{\texttt{#1}}
\expandafter\ifx\csname urlstyle\endcsname\relax
  \providecommand{\doi}[1]{doi: #1}\else
  \providecommand{\doi}{doi: \begingroup \urlstyle{rm}\Url}\fi

\bibitem[Allan and Rugg(1997)]{allan1997event}
Ken Allan and MD Rugg.
\newblock An event-related potential study of explicit memory on tests of cued
  recall and recognition.
\newblock \emph{Neuropsychologia}, 35\penalty0 (4):\penalty0 387--397, 1997.

\bibitem[Banerjee and Lavie(2005)]{banerjee2005meteor}
Satanjeev Banerjee and Alon Lavie.
\newblock Meteor: An automatic metric for mt evaluation with improved
  correlation with human judgments.
\newblock In \emph{Proceedings of the acl workshop on intrinsic and extrinsic
  evaluation measures for machine translation and/or summarization}, pages
  65--72, 2005.

\bibitem[Carion et~al.(2020)Carion, Massa, Synnaeve, Usunier, Kirillov, and
  Zagoruyko]{carion2020end}
Nicolas Carion, Francisco Massa, Gabriel Synnaeve, Nicolas Usunier, Alexander
  Kirillov, and Sergey Zagoruyko.
\newblock End-to-end object detection with transformers.
\newblock In \emph{European conference on computer vision}, pages 213--229.
  Springer, 2020.

\bibitem[Chadha et~al.(2020)Chadha, Arora, and Kaloty]{chadha2020iperceive}
Aman Chadha, Gurneet Arora, and Navpreet Kaloty.
\newblock iperceive: Applying common-sense reasoning to multi-modal dense video
  captioning and video question answering.
\newblock \emph{arXiv preprint arXiv:2011.07735}, 2020.

\bibitem[Chen et~al.(2023)Chen, Pan, Li, Yao, Chao, and Mei]{chen2023retrieval}
Jingwen Chen, Yingwei Pan, Yehao Li, Ting Yao, Hongyang Chao, and Tao Mei.
\newblock Retrieval augmented convolutional encoder-decoder networks for video
  captioning.
\newblock \emph{ACM Transactions on Multimedia Computing, Communications and
  Applications}, 19\penalty0 (1s):\penalty0 1--24, 2023.

\bibitem[Chen and Jiang(2021)]{chen2021towards}
Shaoxiang Chen and Yu-Gang Jiang.
\newblock Towards bridging event captioner and sentence localizer for weakly
  supervised dense event captioning.
\newblock In \emph{Proceedings of the IEEE/CVF Conference on Computer Vision
  and Pattern Recognition}, pages 8425--8435, 2021.

\bibitem[Chen et~al.(2017)Chen, Chen, Jin, and Hauptmann]{chen2017video}
Shizhe Chen, Jia Chen, Qin Jin, and Alexander Hauptmann.
\newblock Video captioning with guidance of multimodal latent topics.
\newblock In \emph{Proceedings of the 25th ACM international conference on
  Multimedia}, pages 1838--1846, 2017.

\bibitem[Cheng et~al.(2023)Cheng, Wang, Lei, Crandall, Bansal, and
  Bertasius]{cheng2023vindlu}
Feng Cheng, Xizi Wang, Jie Lei, David Crandall, Mohit Bansal, and Gedas
  Bertasius.
\newblock Vindlu: A recipe for effective video-and-language pretraining.
\newblock In \emph{Proceedings of the IEEE/CVF Conference on Computer Vision
  and Pattern Recognition}, pages 10739--10750, 2023.

\bibitem[Deng et~al.(2021)Deng, Chen, Chen, He, and Wu]{Deng_2021_CVPR}
Chaorui Deng, Shizhe Chen, Da Chen, Yuan He, and Qi Wu.
\newblock Sketch, ground, and refine: Top-down dense video captioning.
\newblock In \emph{Proceedings of the IEEE/CVF Conference on Computer Vision
  and Pattern Recognition (CVPR)}, pages 234--243, 2021.

\bibitem[Dosovitskiy et~al.(2020)Dosovitskiy, Beyer, Kolesnikov, Weissenborn,
  Zhai, Unterthiner, Dehghani, Minderer, Heigold, Gelly,
  et~al.]{dosovitskiy2020image}
Alexey Dosovitskiy, Lucas Beyer, Alexander Kolesnikov, Dirk Weissenborn,
  Xiaohua Zhai, Thomas Unterthiner, Mostafa Dehghani, Matthias Minderer, Georg
  Heigold, Sylvain Gelly, et~al.
\newblock An image is worth 16x16 words: Transformers for image recognition at
  scale.
\newblock \emph{arXiv preprint arXiv:2010.11929}, 2020.

\bibitem[Fujita et~al.(2020)Fujita, Hirao, Kamigaito, Okumura, and
  Nagata]{fujita2020soda}
Soichiro Fujita, Tsutomu Hirao, Hidetaka Kamigaito, Manabu Okumura, and Masaaki
  Nagata.
\newblock Soda: Story oriented dense video captioning evaluation framework.
\newblock In \emph{Computer Vision--ECCV 2020: 16th European Conference,
  Glasgow, UK, August 23--28, 2020, Proceedings, Part VI 16}, pages 517--531.
  Springer, 2020.

\bibitem[Gao et~al.(2017)Gao, Guo, Zhang, Xu, and Shen]{7984828}
Lianli Gao, Zhao Guo, Hanwang Zhang, Xing Xu, and Heng~Tao Shen.
\newblock Video captioning with attention-based lstm and semantic consistency.
\newblock \emph{IEEE Transactions on Multimedia}, 19\penalty0 (9):\penalty0
  2045--2055, 2017.

\bibitem[Hayes et~al.(2022)Hayes, Zhang, Yin, Pang, Sheng, Yang, Ge, Hu, and
  Parikh]{hayes2022mugen}
Thomas Hayes, Songyang Zhang, Xi Yin, Guan Pang, Sasha Sheng, Harry Yang,
  Songwei Ge, Qiyuan Hu, and Devi Parikh.
\newblock Mugen: A playground for video-audio-text multimodal understanding and
  generation.
\newblock In \emph{European Conference on Computer Vision}, pages 431--449.
  Springer, 2022.

\bibitem[Iashin and Rahtu(2020{\natexlab{a}})]{iashin2020better}
Vladimir Iashin and Esa Rahtu.
\newblock A better use of audio-visual cues: Dense video captioning with
  bi-modal transformer.
\newblock \emph{arXiv preprint arXiv:2005.08271}, 2020{\natexlab{a}}.

\bibitem[Iashin and Rahtu(2020{\natexlab{b}})]{iashin2020multi}
Vladimir Iashin and Esa Rahtu.
\newblock Multi-modal dense video captioning.
\newblock In \emph{Proceedings of the IEEE/CVF conference on computer vision
  and pattern recognition workshops}, pages 958--959, 2020{\natexlab{b}}.

\bibitem[Jing et~al.(2023)Jing, Zhang, Zeng, Gao, Song, and Shen]{10183355}
Shuaiqi Jing, Haonan Zhang, Pengpeng Zeng, Lianli Gao, Jingkuan Song, and
  Heng~Tao Shen.
\newblock Memory-based augmentation network for video captioning.
\newblock \emph{IEEE Transactions on Multimedia}, pages 1--13, 2023.

\bibitem[Krishna et~al.(2017)Krishna, Hata, Ren, Fei-Fei, and
  Carlos~Niebles]{krishna2017dense}
Ranjay Krishna, Kenji Hata, Frederic Ren, Li Fei-Fei, and Juan Carlos~Niebles.
\newblock Dense-captioning events in videos.
\newblock In \emph{Proceedings of the IEEE international conference on computer
  vision}, pages 706--715, 2017.

\bibitem[Lewis et~al.(2020)Lewis, Perez, Piktus, Petroni, Karpukhin, Goyal,
  K{\"u}ttler, Lewis, Yih, Rockt{\"a}schel, et~al.]{lewis2020retrieval}
Patrick Lewis, Ethan Perez, Aleksandra Piktus, Fabio Petroni, Vladimir
  Karpukhin, Naman Goyal, Heinrich K{\"u}ttler, Mike Lewis, Wen-tau Yih, Tim
  Rockt{\"a}schel, et~al.
\newblock Retrieval-augmented generation for knowledge-intensive nlp tasks.
\newblock \emph{Advances in Neural Information Processing Systems},
  33:\penalty0 9459--9474, 2020.

\bibitem[Li et~al.(2022)Li, Li, Xiong, and Hoi]{li2022blip}
Junnan Li, Dongxu Li, Caiming Xiong, and Steven Hoi.
\newblock Blip: Bootstrapping language-image pre-training for unified
  vision-language understanding and generation.
\newblock In \emph{International Conference on Machine Learning}, pages
  12888--12900. PMLR, 2022.

\bibitem[Li et~al.(2018)Li, Yao, Pan, Chao, and Mei]{li2018jointly}
Yehao Li, Ting Yao, Yingwei Pan, Hongyang Chao, and Tao Mei.
\newblock Jointly localizing and describing events for dense video captioning.
\newblock In \emph{Proceedings of the IEEE conference on computer vision and
  pattern recognition}, pages 7492--7500, 2018.

\bibitem[Lin et~al.(2022)Lin, Li, Lin, Ahmed, Gan, Liu, Lu, and
  Wang]{lin2022swinbert}
Kevin Lin, Linjie Li, Chung-Ching Lin, Faisal Ahmed, Zhe Gan, Zicheng Liu,
  Yumao Lu, and Lijuan Wang.
\newblock Swinbert: End-to-end transformers with sparse attention for video
  captioning.
\newblock In \emph{Proceedings of the IEEE/CVF Conference on Computer Vision
  and Pattern Recognition}, pages 17949--17958, 2022.

\bibitem[Luo et~al.(2020)Luo, Ji, Shi, Huang, Duan, Li, Li, Bharti, and
  Zhou]{luo2020univl}
Huaishao Luo, Lei Ji, Botian Shi, Haoyang Huang, Nan Duan, Tianrui Li, Jason
  Li, Taroon Bharti, and Ming Zhou.
\newblock Univl: A unified video and language pre-training model for multimodal
  understanding and generation, 2020.

\bibitem[Mun et~al.(2019)Mun, Yang, Ren, Xu, and Han]{mun2019streamlined}
Jonghwan Mun, Linjie Yang, Zhou Ren, Ning Xu, and Bohyung Han.
\newblock Streamlined dense video captioning.
\newblock In \emph{Proceedings of the IEEE/CVF conference on computer vision
  and pattern recognition}, pages 6588--6597, 2019.

\bibitem[Pan et~al.(2016)Pan, Mei, Yao, Li, and Rui]{pan2016jointly}
Yingwei Pan, Tao Mei, Ting Yao, Houqiang Li, and Yong Rui.
\newblock Jointly modeling embedding and translation to bridge video and
  language.
\newblock In \emph{Proceedings of the IEEE conference on computer vision and
  pattern recognition}, pages 4594--4602, 2016.

\bibitem[Papineni et~al.(2002)Papineni, Roukos, Ward, and
  Zhu]{papineni2002bleu}
Kishore Papineni, Salim Roukos, Todd Ward, and Wei-Jing Zhu.
\newblock Bleu: a method for automatic evaluation of machine translation.
\newblock In \emph{Proceedings of the 40th annual meeting of the Association
  for Computational Linguistics}, pages 311--318, 2002.

\bibitem[Pei et~al.(2019)Pei, Zhang, Wang, Ke, Shen, and Tai]{pei2019memory}
Wenjie Pei, Jiyuan Zhang, Xiangrong Wang, Lei Ke, Xiaoyong Shen, and Yu-Wing
  Tai.
\newblock Memory-attended recurrent network for video captioning.
\newblock In \emph{Proceedings of the IEEE/CVF Conference on Computer Vision
  and Pattern Recognition}, pages 8347--8356, 2019.

\bibitem[Qi et~al.(2019)Qi, Wang, Li, and Luo]{qi2019sports}
Mengshi Qi, Yunhong Wang, Annan Li, and Jiebo Luo.
\newblock Sports video captioning via attentive motion representation and group
  relationship modeling.
\newblock \emph{IEEE Transactions on Circuits and Systems for Video
  Technology}, 30\penalty0 (8):\penalty0 2617--2633, 2019.

\bibitem[Radford et~al.(2021)Radford, Kim, Hallacy, Ramesh, Goh, Agarwal,
  Sastry, Askell, Mishkin, Clark, et~al.]{radford2021learning}
Alec Radford, Jong~Wook Kim, Chris Hallacy, Aditya Ramesh, Gabriel Goh,
  Sandhini Agarwal, Girish Sastry, Amanda Askell, Pamela Mishkin, Jack Clark,
  et~al.
\newblock Learning transferable visual models from natural language
  supervision.
\newblock In \emph{International conference on machine learning}, pages
  8748--8763. PMLR, 2021.

\bibitem[Rahman et~al.(2019)Rahman, Xu, and Sigal]{rahman2019watch}
Tanzila Rahman, Bicheng Xu, and Leonid Sigal.
\newblock Watch, listen and tell: Multi-modal weakly supervised dense event
  captioning.
\newblock In \emph{Proceedings of the IEEE/CVF international conference on
  computer vision}, pages 8908--8917, 2019.

\bibitem[Ramos et~al.(2023)Ramos, Elliott, and Martins]{ramos2023retrieval}
Rita Ramos, Desmond Elliott, and Bruno Martins.
\newblock Retrieval-augmented image captioning.
\newblock \emph{arXiv preprint arXiv:2302.08268}, 2023.

\bibitem[Ramos et~al.(2021)Ramos, Pereira, Moniz, Carvalho, and
  Martins]{ramos2021retrieval}
Rita~Parada Ramos, Patr{\'\i}cia Pereira, Helena Moniz, Joao~Paulo Carvalho,
  and Bruno Martins.
\newblock Retrieval augmentation for deep neural networks.
\newblock In \emph{2021 International Joint Conference on Neural Networks
  (IJCNN)}, pages 1--8. IEEE, 2021.

\bibitem[Rohrbach et~al.(2013)Rohrbach, Qiu, Titov, Thater, Pinkal, and
  Schiele]{rohrbach2013translating}
Marcus Rohrbach, Wei Qiu, Ivan Titov, Stefan Thater, Manfred Pinkal, and Bernt
  Schiele.
\newblock Translating video content to natural language descriptions.
\newblock In \emph{Proceedings of the IEEE international conference on computer
  vision}, pages 433--440, 2013.

\bibitem[Rugg et~al.(1998)Rugg, Fletcher, Allan, Frith, Frackowiak, and
  Dolan]{rugg1998neural}
Michael~D Rugg, Paul~C Fletcher, Kevin Allan, Chris~D Frith, RSJ Frackowiak,
  and Raymond~J Dolan.
\newblock Neural correlates of memory retrieval during recognition memory and
  cued recall.
\newblock \emph{Neuroimage}, 8\penalty0 (3):\penalty0 262--273, 1998.

\bibitem[Saharia et~al.(2022)Saharia, Chan, Saxena, Li, Whang, Denton,
  Ghasemipour, Gontijo~Lopes, Karagol~Ayan, Salimans,
  et~al.]{saharia2022photorealistic}
Chitwan Saharia, William Chan, Saurabh Saxena, Lala Li, Jay Whang, Emily~L
  Denton, Kamyar Ghasemipour, Raphael Gontijo~Lopes, Burcu Karagol~Ayan, Tim
  Salimans, et~al.
\newblock Photorealistic text-to-image diffusion models with deep language
  understanding.
\newblock \emph{Advances in Neural Information Processing Systems},
  35:\penalty0 36479--36494, 2022.

\bibitem[Sarto et~al.(2022)Sarto, Cornia, Baraldi, and
  Cucchiara]{sarto2022retrieval}
Sara Sarto, Marcella Cornia, Lorenzo Baraldi, and Rita Cucchiara.
\newblock Retrieval-augmented transformer for image captioning.
\newblock In \emph{Proceedings of the 19th International Conference on
  Content-based Multimedia Indexing}, pages 1--7, 2022.

\bibitem[Seo et~al.(2022)Seo, Nagrani, Arnab, and Schmid]{seo2022end}
Paul~Hongsuck Seo, Arsha Nagrani, Anurag Arnab, and Cordelia Schmid.
\newblock End-to-end generative pretraining for multimodal video captioning.
\newblock In \emph{Proceedings of the IEEE/CVF Conference on Computer Vision
  and Pattern Recognition}, pages 17959--17968, 2022.

\bibitem[Shen et~al.(2017)Shen, Li, Su, Li, Chen, Jiang, and
  Xue]{shen2017weakly}
Zhiqiang Shen, Jianguo Li, Zhou Su, Minjun Li, Yurong Chen, Yu-Gang Jiang, and
  Xiangyang Xue.
\newblock Weakly supervised dense video captioning.
\newblock In \emph{Proceedings of the IEEE Conference on Computer Vision and
  Pattern Recognition}, pages 1916--1924, 2017.

\bibitem[Shi et~al.(2019)Shi, Ji, Liang, Duan, Chen, Niu, and
  Zhou]{shi2019dense}
Botian Shi, Lei Ji, Yaobo Liang, Nan Duan, Peng Chen, Zhendong Niu, and Ming
  Zhou.
\newblock Dense procedure captioning in narrated instructional videos.
\newblock In \emph{Proceedings of the 57th annual meeting of the association
  for computational linguistics}, pages 6382--6391, 2019.

\bibitem[Vedantam et~al.(2015)Vedantam, Lawrence~Zitnick, and
  Parikh]{vedantam2015cider}
Ramakrishna Vedantam, C Lawrence~Zitnick, and Devi Parikh.
\newblock Cider: Consensus-based image description evaluation.
\newblock In \emph{Proceedings of the IEEE conference on computer vision and
  pattern recognition}, pages 4566--4575, 2015.

\bibitem[Venugopalan et~al.(2014)Venugopalan, Xu, Donahue, Rohrbach, Mooney,
  and Saenko]{venugopalan2014translating}
Subhashini Venugopalan, Huijuan Xu, Jeff Donahue, Marcus Rohrbach, Raymond
  Mooney, and Kate Saenko.
\newblock Translating videos to natural language using deep recurrent neural
  networks.
\newblock \emph{arXiv preprint arXiv:1412.4729}, 2014.

\bibitem[Venugopalan et~al.(2015)Venugopalan, Rohrbach, Donahue, Mooney,
  Darrell, and Saenko]{venugopalan2015sequence}
Subhashini Venugopalan, Marcus Rohrbach, Jeffrey Donahue, Raymond Mooney,
  Trevor Darrell, and Kate Saenko.
\newblock Sequence to sequence-video to text.
\newblock In \emph{Proceedings of the IEEE international conference on computer
  vision}, pages 4534--4542, 2015.

\bibitem[Wang et~al.(2018{\natexlab{a}})Wang, Ma, Zhang, and
  Liu]{wang2018reconstruction}
Bairui Wang, Lin Ma, Wei Zhang, and Wei Liu.
\newblock Reconstruction network for video captioning.
\newblock In \emph{Proceedings of the IEEE conference on computer vision and
  pattern recognition}, pages 7622--7631, 2018{\natexlab{a}}.

\bibitem[Wang et~al.(2018{\natexlab{b}})Wang, Jiang, Ma, Liu, and
  Xu]{wang2018bidirectional}
Jingwen Wang, Wenhao Jiang, Lin Ma, Wei Liu, and Yong Xu.
\newblock Bidirectional attentive fusion with context gating for dense video
  captioning.
\newblock In \emph{Proceedings of the IEEE conference on computer vision and
  pattern recognition}, pages 7190--7198, 2018{\natexlab{b}}.

\bibitem[Wang et~al.(2020{\natexlab{a}})Wang, Zheng, and Yu]{wang2020dense}
Teng Wang, Huicheng Zheng, and Mingjing Yu.
\newblock Dense-captioning events in videos: Sysu submission to activitynet
  challenge 2020.
\newblock \emph{arXiv preprint arXiv:2006.11693}, 2020{\natexlab{a}}.

\bibitem[Wang et~al.(2020{\natexlab{b}})Wang, Zheng, Yu, Tian, and
  Hu]{wang2020event}
Teng Wang, Huicheng Zheng, Mingjing Yu, Qian Tian, and Haifeng Hu.
\newblock Event-centric hierarchical representation for dense video captioning.
\newblock \emph{IEEE Transactions on Circuits and Systems for Video
  Technology}, 31\penalty0 (5):\penalty0 1890--1900, 2020{\natexlab{b}}.

\bibitem[Wang et~al.(2021)Wang, Zhang, Lu, Zheng, Cheng, and Luo]{wang2021end}
Teng Wang, Ruimao Zhang, Zhichao Lu, Feng Zheng, Ran Cheng, and Ping Luo.
\newblock End-to-end dense video captioning with parallel decoding.
\newblock In \emph{Proceedings of the IEEE/CVF International Conference on
  Computer Vision}, pages 6847--6857, 2021.

\bibitem[Xu et~al.(2019)Xu, Zhao, Yang, Ao, Cheng, and Tian]{xu2019unified}
Chunpu Xu, Wei Zhao, Min Yang, Xiang Ao, Wangrong Cheng, and Jinwen Tian.
\newblock A unified generation-retrieval framework for image captioning.
\newblock In \emph{Proceedings of the 28th ACM International Conference on
  Information and Knowledge Management}, pages 2313--2316, 2019.

\bibitem[Yang et~al.(2023)Yang, Nagrani, Seo, Miech, Pont-Tuset, Laptev, Sivic,
  and Schmid]{yang2023vid2seq}
Antoine Yang, Arsha Nagrani, Paul~Hongsuck Seo, Antoine Miech, Jordi
  Pont-Tuset, Ivan Laptev, Josef Sivic, and Cordelia Schmid.
\newblock Vid2seq: Large-scale pretraining of a visual language model for dense
  video captioning.
\newblock In \emph{CVPR}, 2023.

\bibitem[Yang and Yuan(2018)]{yang2018hierarchical}
Dali Yang and Chun Yuan.
\newblock Hierarchical context encoding for events captioning in videos.
\newblock In \emph{2018 25th IEEE International Conference on Image Processing
  (ICIP)}, pages 1288--1292. IEEE, 2018.

\bibitem[Zhang et~al.(2022)Zhang, Song, and Jin]{zhang2022unifying}
Qi Zhang, Yuqing Song, and Qin Jin.
\newblock Unifying event detection and captioning as sequence generation via
  pre-training.
\newblock In \emph{European Conference on Computer Vision}, pages 363--379.
  Springer, 2022.

\bibitem[Zhang et~al.(2021)Zhang, Qi, Yuan, Shan, Li, Deng, and
  Hu]{zhang2021open}
Ziqi Zhang, Zhongang Qi, Chunfeng Yuan, Ying Shan, Bing Li, Ying Deng, and
  Weiming Hu.
\newblock Open-book video captioning with retrieve-copy-generate network.
\newblock In \emph{Proceedings of the IEEE/CVF conference on computer vision
  and pattern recognition}, pages 9837--9846, 2021.

\bibitem[Zhao et~al.(2020)Zhao, Li, Peng, Yang, and Zhang]{zhao2020image}
Shanshan Zhao, Lixiang Li, Haipeng Peng, Zihang Yang, and Jiaxuan Zhang.
\newblock Image caption generation via unified retrieval and generation-based
  method.
\newblock \emph{Applied Sciences}, 10\penalty0 (18):\penalty0 6235, 2020.

\bibitem[Zhou et~al.(2018{\natexlab{a}})Zhou, Xu, and Corso]{zhou2018towards}
Luowei Zhou, Chenliang Xu, and Jason Corso.
\newblock Towards automatic learning of procedures from web instructional
  videos.
\newblock In \emph{Proceedings of the AAAI Conference on Artificial
  Intelligence}, 2018{\natexlab{a}}.

\bibitem[Zhou et~al.(2018{\natexlab{b}})Zhou, Zhou, Corso, Socher, and
  Xiong]{zhou2018end}
Luowei Zhou, Yingbo Zhou, Jason~J Corso, Richard Socher, and Caiming Xiong.
\newblock End-to-end dense video captioning with masked transformer.
\newblock In \emph{Proceedings of the IEEE conference on computer vision and
  pattern recognition}, pages 8739--8748, 2018{\natexlab{b}}.

\bibitem[Zhu et~al.(2022)Zhu, Pang, Thapliyal, Wang, and Soricut]{zhu2022end}
Wanrong Zhu, Bo Pang, Ashish~V Thapliyal, William~Yang Wang, and Radu Soricut.
\newblock End-to-end dense video captioning as sequence generation.
\newblock \emph{International Conference on Computational Linguistics
  (COLING)}, 2022.

\bibitem[Zhu et~al.(2020)Zhu, Su, Lu, Li, Wang, and Dai]{zhu2020deformable}
Xizhou Zhu, Weijie Su, Lewei Lu, Bin Li, Xiaogang Wang, and Jifeng Dai.
\newblock Deformable detr: Deformable transformers for end-to-end object
  detection.
\newblock \emph{arXiv preprint arXiv:2010.04159}, 2020.

\end{thebibliography}
}
\clearpage
\setcounter{page}{1}
\maketitlesupplementary
\begin{appendices}

\begin{table}[!t]
 \caption{\textbf{Ablation study to verify the effect of cross-attention design in the versatile decoder.} VC$\rightarrow$TC denotes the structure where textual cross-attention follows after visual cross-attention (Figure 2 (c)). TC$\rightarrow$VC denotes the structure that is implemented by making visual cross-attention follow after textual cross-attention. Parallel cross-attention is implemented by merging each cross-attention output. The performance is measured in YouCook2.}
\centering
\resizebox{1\columnwidth}{!}{
    \begin{tabular}{@{}c|ccccc@{}}
    \toprule
     Cross-Attention Design & CIDEr & METEOR
     &  SODA$\_c$ &F1 \\
    \midrule
    Sequential (VC$\rightarrow$TC) &31.66 & \textbf{6.08} 
    & \textbf{5.34} & \textbf{28.43}\\
    Sequential (TC$\rightarrow$VC) &\textbf{33.33} & 5.77
    & 5.14 & 27.03\\
    Parallel cross-attention& 30.63& 5.55&5.13 & 26.96\\
    \bottomrule
    \end{tabular}
}
\label{tab:ca_order}
\end{table}

\section{Additional Analysis of CM$^2$}

\textbf{Cross-attention Design.}
We explore the effect of the order of constructing cross-attention modules in the versatile decoder on the model's localization and caption generation abilities. VC$\rightarrow$TC is the method used in this study where visual features are incorporated first with visual cross-attention and textual cross-attention follows after visual cross-attention. TC$\rightarrow$VC is designed by making temporal cross-attention first, then visual cross-attention follows. In Table \ref{tab:ca_order}, it is observed that VC$\rightarrow$TC shows better performance compared to TC$\rightarrow$VC on METEOR, SODA$\_c$, and F1. We also conducted an experiment   of the Parallel cross-attention. Parallel cross-attention is implemented by conducting cross-attention separately for VC and TC, and then merging them. Compared to parallel cross-attention, sequential cross-attention demonstrates better performance in both localization and captioning tasks overall.

\begin{table}[t!]
\caption{\textbf{Analysis of different aggregation methods of memory read module in YouCook2 dataset.} }
\centering
\resizebox{\linewidth}{!}{
    \begin{tabular}{@{}l|ccccc@{}}
    \toprule
    Aggregation Type  &  CIDEr & METEOR &  BLEU4 & SODA$\_c$& F1 \\
    \midrule
    Attention & 32.21 & 5.84 & 1.90& 5.19 & 27.05\\
    Average Pooling  & 31.66 &6.08 &1.63& 5.34& 28.40 \\
    \bottomrule
    \end{tabular}
}
\vspace{-0.2cm}
\label{tab:ret_aggregation}
\end{table}
 
\noindent\textbf{Effect of Selected Features Aggregation Methods.} In \tabref{ret_aggregation}, we explore methods for aggregating retrieved text features as segment-level semantic information. We compare two different aggregation methods. The attention method shows fairly comparable results in event captioning but exhibited relatively lower performance in event localization. On average pooling method, as outlined in Section \ref{sec:MemRetrieval}, showed consistently comparable performance in both event captioning and event localization. 

\noindent\textbf{Memory Bank Size.}
We further analyze our model with respect to the memory bank size. To investigate the effect of memory size on our model, we randomly sample text features to construct external memory. For example, the memory size 10\% is implemented by sampling 10\% of training data captions to construct the external memory, and the inference is conducted with the reduced memory. To reduce the effect of randomness, we report average scores calculated from 50 repetitions with different memory sampling, excluding cases where the memory size ratio is 0\% or 100\%. We investigate the effect of memory size on a log scale. As shown in Table~\ref{tab:mem_size}, as the amount of information in the memory increases, CIDEr and METEOR improve. With a small memory size, our method could achieve higher performance compared with the model with no memory (no memory is implemented by replacing text features in the external memory with zero features).

\begin{table}[t!]
\caption{\textbf{Analysis of our model with respect to the memory bank size in YouCook2 dataset. } The performance is measured by changing the memory bank size. The memory ratio means that we randomly sample the sentence features from the training data to construct text features in the external memory. The average scores are calculated from 50 repetitions with different memory sampling}  
\centering
\scalebox{0.9}{
\begin{tabular}{@{}c|cccc@{}}
\toprule

Memory Ratio&CIDEr & METEOR &BLEU4& SODA$\_c$ \\
\midrule
0\% &27.91&5.66&1.24&4.92\\
0.01\% &29.59&5.74&1.53&5.09\\
0.1\% &30.70&5.75&1.67&5.33\\
1\% &31.17&5.79&1.69&5.39\\
10\% &31.34&5.92&1.67&5.37\\
100\% &31.66 & 6.08 &1.63& 5.34  \\
\bottomrule
\end{tabular}
}
\label{tab:mem_size}
\end{table}

\begin{table}[t!]
\caption{\textbf{Effect of memory bank construction.} YC2, ANet, Epic, MSR denote YouCook2, ActivityNet captions, Epic Kitchens, and MSR-VTT datasets. For memory bank, only training set is used.}
\vspace{-0.2cm}
\centering
\resizebox{0.9\columnwidth}{!}{
    \begin{tabular}{@{}c|c|ccccc@{}}
    \toprule
    Test& Memory Bank & CIDEr & METEOR 
     &  SODA$\_c$ &F1 \\
    \midrule
    \multirow{4}{*}{YC2} &YC2&31.66 & 6.08
    & 5.34 & 28.43\\
    &YC2+ANet&31.66 & 6.01
    & 5.38 & 28.53\\
    &Epic & 31.58 & 5.91 &5.36 &28.51\\
    &YC2+Epic & 32.21 & 6.08 &5.39 &28.50\\
    \midrule
    \multirow{4}{*}{ANet} &ANet & 33.01 & 8.55 &6.18&55.21\\
    &ANet+YC2&33.13 &8.57& 6.19 & 55.18\\
    &MSR & 33.09&8.55 &6.12&54.94\\
    &ANet+MSR &33.34&8.60&6.19&55.18\\
    \bottomrule
    \end{tabular}
}
\label{tab:mem_design}
\end{table}

\noindent\textbf{Scalability of Memory Bank.}
We show additional results in Table~\ref{tab:mem_design} below when the memory is built by combining training sets of YouCook2 and ActivityNet Captions as a unified external knowledge. The unified memory still shows comparable performance. Moreover, we further present experimental results of constructing the memory with another dataset. When the memory is built with the dataset from the same domain (e.g., YC2+Epic for YC2, ANet+MSR for ANet), we observe performance improvement without additional training. Note that our method builds the memory with CLIP text features from caption datasets, which increases the scalability of the method.

\noindent\section{Performance of Paragraph Captioning}
We further present the results of our model in terms of paragraph captioning. Note that any additional training is not conducted for paragraph captioning. We just measure the performance of our model by collecting generated captions in order and calculating the performance for the query video at a paragraph level. Table \ref{tab:para} shows the results of the models at a paragraph level. As shown in the table, Vid2seq~\cite{yang2023vid2seq}, which utilizes an additional 15 million videos for pre-training, achieves the best performance. Our method shows comparable performance without pre-train with extra videos. In our future work, we plan to enhance paragraph generation by incorporating optimized sentence retrieval and training schemes specifically tailored for paragraph generation.

\begin{table}[t!]
\caption{\textbf{Performance of Paragraph Captioning in ActivityNet.} Bold means the highest score. Underline means 2nd score. \# PT denotes the number of videos used for pre-training. {$^{\dagger}$} denotes results reproduced from official implementation in our environment.
}
\centering
\resizebox{0.9\linewidth}{!}{
    \begin{tabular}{@{}l|l|c|cc@{}}
    \toprule
    \multirow{2}{*}{Method} & \multirow{2}{*}{Backbone}&\multirow{2}{*}{\#PT}&\multicolumn{2}{c}{ActivityNet (val-ae)} \\
    & && CIDEr & METEOR  \\
    \midrule
    Vid2Seq\cite{yang2023vid2seq} & CLIP & 15M & \textbf{28.00} & \textbf{17.00}\\
    \midrule
    PDVC\cite{wang2021end} & TSN &- &20.50 & 15.80 \\
    PDVC{$^{\dagger}$}\cite{wang2021end} & CLIP &- &23.74 & 16.03 \\
    \textbf{Ours} & CLIP &-&\underline{25.31} & \underline{16.47} \\
    \bottomrule
    \end{tabular}
}
\label{tab:para}
\end{table}

\begin{figure*}[t]
    \centering
    \includegraphics[width=0.88\linewidth]{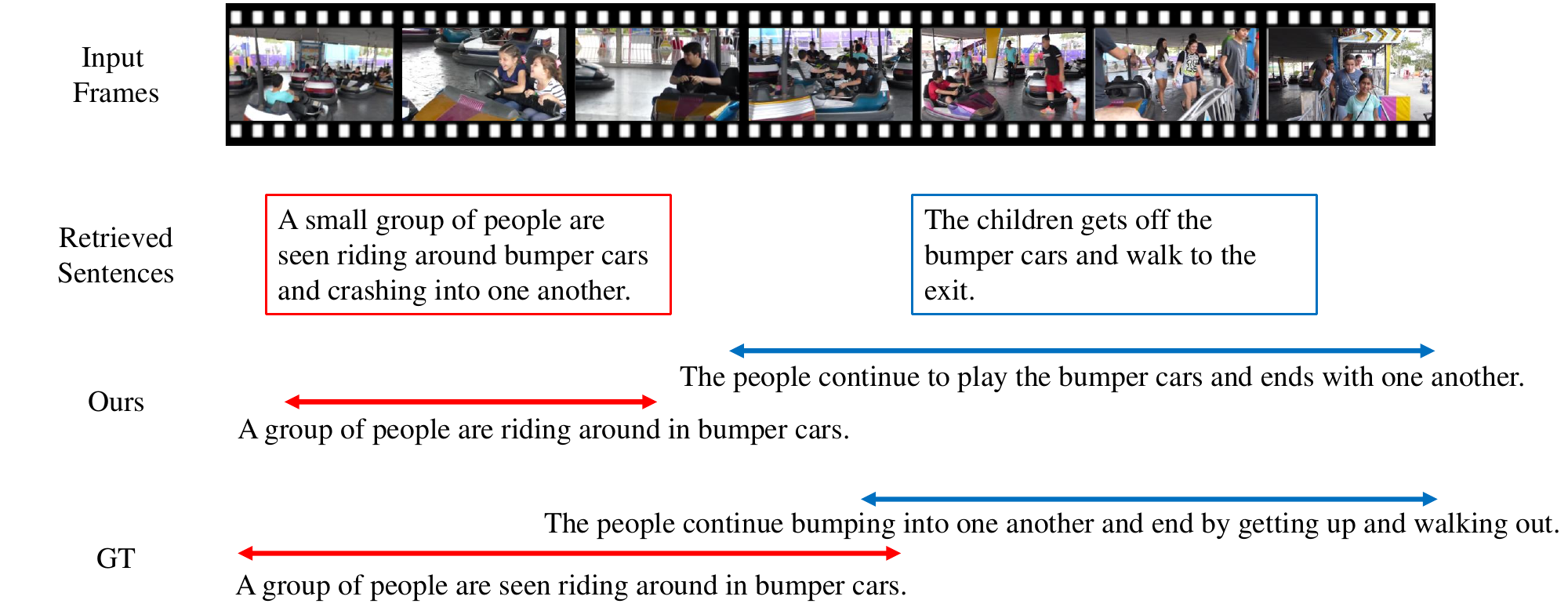}
    \caption{\textbf{Example of predictions from our method on ActivityNet Captions dataset.} We show a comparison with the ground truth. Retrieved sentences are example results from retrieval that have the highest similarity to the corresponding segments of input frames. Each retrieved sentence is utilized in our model's predictions for the segments with the corresponding color.}
    \label{fig:supple_qualitative1}
\end{figure*}

\section{Qualitative Results}

\begin{figure*}[t]
    \centering
    \includegraphics[width=0.88\linewidth]{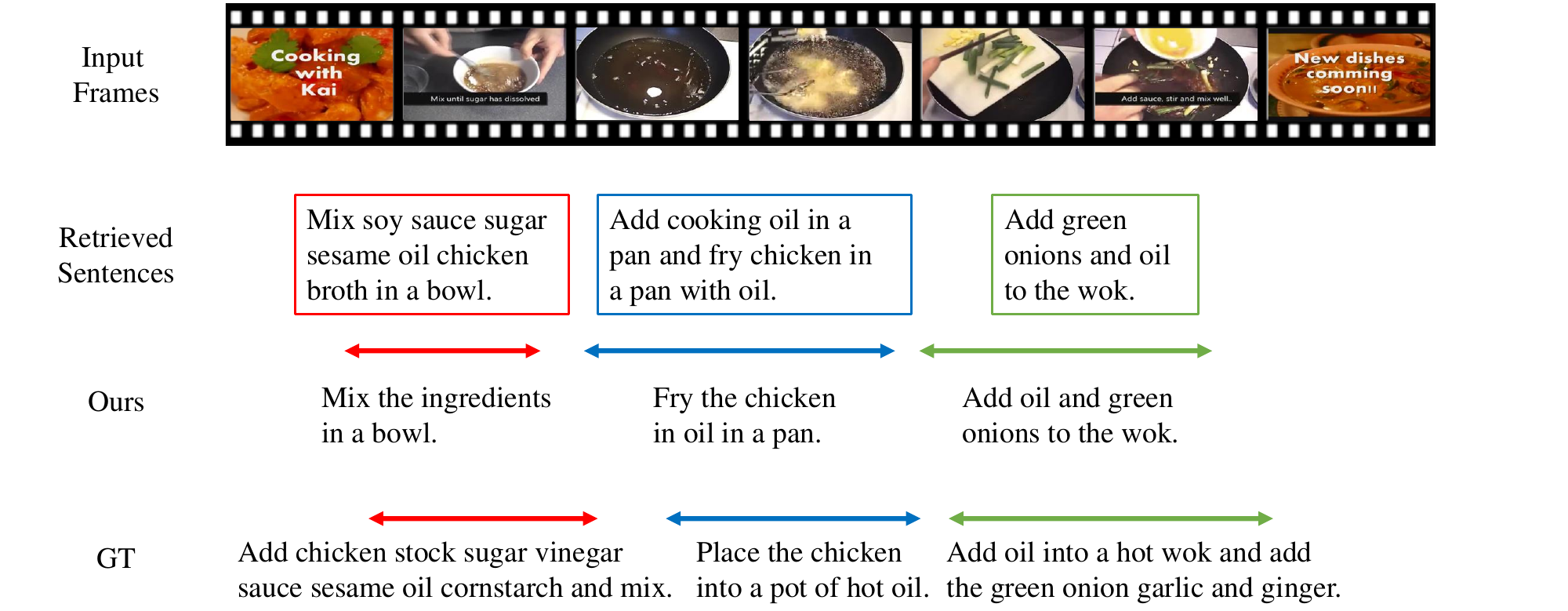}
    \caption{\textbf{Example of predictions from our method on YouCook2 dataset.} We show a comparison with the ground truth. Retrieved sentences are example results from retrieval that have the highest semantic similarity to the corresponding segments of input frames. Each retrieved sentence is utilized in our model's predictions for the segments with the corresponding color.}
    \label{fig:supple_qualitative2}
\end{figure*}

In Figure~\ref{fig:supple_qualitative1} and Figure~\ref{fig:supple_qualitative2}, we show additional qualitative examples of our approach. As shown in the figures, memory retrieval could provide relevant semantics for analyzing input query video. As a result, our approach could yield precise event boundaries and captions. The semantic information retrieved from memory assists in semantic predictions during the caption generation process as shown Section 4.

\end{appendices}

\end{document}